\documentclass[11pt]{article}

\usepackage[preprint]{acl}

\usepackage{times}
\usepackage{latexsym}

\usepackage[T1]{fontenc}

\usepackage[utf8]{inputenc}

\usepackage{microtype}

\usepackage{inconsolata}

\usepackage{graphicx}

\title{No One-Size-Fits-All: Building Systems For Translation to Bashkir, Kazakh, Kyrgyz, Tatar and Chuvash Using Synthetic And Original Data}

\author{Dmitry Karpov \\
  PAO Severstal / Moscow, Russia \\
  \texttt{dimakarp1996@yandex.ru} \\}

\begin{document}
\maketitle
\begin{abstract}
We explore machine translation for five Turkic language pairs: Russian-Bashkir, Russian-Kazakh, Russian-Kyrgyz, English-Tatar, English-Chuvash. Fine-tuning nllb-200-distilled-600M with LoRA on synthetic data achieved chrF++ 49.71 for Kazakh and 46.94 for Bashkir. Prompting DeepSeek-V3.2 with retrieved similar examples achieved chrF++ 39.47 for Chuvash. For Tatar, zero-shot or retrieval-based approaches achieved chrF++ 41.6, while for Kyrgyz the zero-shot approach reached 45.6. We release the dataset and the obtained weights.
\end{abstract}

\section{Introduction}
Machine translation for low-resource languages remains challenging. The "Machine Translation for Low-Resource Turkic Languages" competition focused on five pairs: Russian-Kazakh, Russian-Kyrgyz, Russian-Bashkir, English-Tatar, English-Chuvash. We investigate multiple approaches to improve translation quality in data-scarce conditions.
\section{Making The Data}
Unfortunately, only a limited amount of high-quality parallel data was available at the time of this study. Therefore, we used a variety of datasets. Specifically, we used the parallel English-Chuvash corpus from \citet{plotnikov2024opendatachuvashdatasets} for translations from English to Chuvash, \citet{opus_100} for English-Tatar, English-Kyrgyz, and English-Kazakh translations, \citet{flores_plus} to obtain the parallel data for Russian, English, Bashkir, Tatar, Kazakh and Kyrgyz, \citet{ipsan_tt} for translations from Russian to Tatar, \citet{global_mmlu}  for the parallel Russian-Kyrgyz data, \cite{tatoeba} for the parallel data for all 5 language pairs, \citet{bash_rus} for the parallel Russian-Bashkir data, \citet{kazparc} and \citet{kazakh_literature_corpus}  for the Russian-Kazakh data and \citet{gourmet} for  Russian-Kyrgyz pair. Unfortunately, we could not use the parallel corpora from the TurkLang-7 project~\citet{turklang} as this corpora was not available at the time of study. The original training set sizes were: 1,190,773 samples for Russian-Bashkir (9,997 in the validation set), 35,429 samples for Russian-Kyrgyz ( 4,845 in the validation set), 367,095 samples for Russian-Kazakh (9,845 in the validation set), 106,777 samples for English-Tatar (3,786 in the validation set), and 193,826 samples for English-Chuvash (9,953 in the validation set).  We report only the official validation scores from the competition leaderboard.

We augmented the original data with synthetic translations from Yandex.Translate\cite{yandextranslate}. To obtain these translations, we translated English phrases into Russian (for pairs where Russian was not the source language) and then translated every Russian phrase to those Turkic languages. We used the "document translation" feature, processing the data in chunks of 50,000 to 200,000 samples due to its large volume. After obtaining synthetic data, we meticulously filtered out any English or Russian phrases from Yandex.Translate that appeared in the test set for any language. At this stage, we also included Russian-Tatar, English-Kazakh, and English-Kyrgyz data.
Data pseudolabeling has long been proven to improve the performance of Transformer-based language models~\cite{pseudolabel}. In our case, pseudolabeling also proved beneficial, as multilingual training without pseudolabeling yielded inferior results in the preliminary experiments.

After augmentation, the training data size increased to 2,457,344 samples for each language pair. 

In addition to these data, we also used translations of MASSIVE dataset \cite{massive} from English to Tatar, Chuvash and Russian, and from Russian to the Bashkir, Kyrgyz and Kazakh (16507 samples). All translations were obtained similarly using Yandex.Translate. However, these translations were obtained later and thus were not used for training the NLLB model. They were used only in the prompting-based solutions.

For the prompting-based approach to English-Chuvash, we additionally obtained translations from Chuvash to English for two other alexantonov corpora: alexantonov\/chuvash\_russian\_parallel and alexantonov\/chuvash\_mono (\citet{plotnikov2024opendatachuvashdatasets}) .  After adding these data to the previously obtained Chuvash-English pairs, the English-Chuvash dataset increased its size to 6.7 million pairs. All this data was also translated to Tatar via Yandex.Translate, and Tatar translations are also provided. However, this data was NOT used in the submissions for Tatar, as a) they were obtained after the training experiments b) building an index from them (see sections below) resulted in the inferior results on the preliminary experiment.

\textbf{We release the resulting dataset YaTURK-7lang, translated into the six languages, here \url{https://huggingface.co/datasets/dimakarp1996/YaTURK-7lang}}. Data used only in the Chuvash solution are marked with the attribute \texttt{only\_index1} set to 0.

\section{Kazakh and Bashkir: Where LoRA And Knowledge Transfer Shined}
We chose \textit{facebook/nllb-200-distilled-600M}~\cite{nllb} as the base model for finetuning. The data for finetuning was preprocessed as follows: additional language tokens for each target language and language pair (ten tokens in total) were added into the tokenizer. Thus, the model input consisted of the prefix of the language pair (e.g. \texttt{<prefix\_rus\_bash>}) with tokenized source language text, and trained the model to predict the target language text, starting its output from the token of the target language (e.g. \texttt{<prefix\_bash>})

We explored 2 main modes of finetuning the model. In the first mode, the model was finetuned for 2 epochs on the data from every language, separately, In the second mode, the model was first finetuned for one epoch on the data from all languages, and then we trained LoRA adapter for every separate language. Neither using LoRA nor finetuning for more than two epochs improved the results.

Specifically, for training adapters, we used DORA\cite{dora}. DORA is the extension of LoRA\cite{LoRA} parameter-efficient finetuning approach. The DORA config was: r=64, alpha=64, LoRA dropout=0.2, PiSSA weight initialization\cite{pissa},target\_modules: q\_proj, v\_proj, k\_proj, out\_proj, fc1, fc2, and shared.  DORA was finetuned with the paged AdamW-8bit\cite{adamw}\cite{8bit} optimizer, with starting learning rate 5e-4 and weight decay 1e-2, train batch size 16 and 8 gradient accumulation steps, linear learning rate scheduler.
For full finetuning, we used the following hyperparameters: batch size 64, 32 gradient accumulation steps, learning rate 2e-4, weight decay 1e-2, 600 warming steps per epoch, paged AdamW-8bit optimizer, cosine learning rate scheduling. In both cases, the maximum sequence length was 128 tokens, and the optimizer state was reset every epoch.
For all models, to obtain generation, we used the following generation settings: min\_length=3, max\_length=150, repetition penalty 1.5, 5 beams.

As one can see from Table\ref{tab:finetune} , the approach of training the model on multiple languages and then finetuning using LoRA outperformed the single-task finetuning, which suggests that knowledge transfer occurs between tasks. Multi-task knowledge transfer has been studied for a long time~\cite{rumtl}. As the Turkic languages in this study are similar, knowledge acquired for one language can help improve performance on another.

\begin{table}
 \caption{Validation set chrF++ (from the competition server) of the NLLB model. Mult-1 means the results of the model finetuned on 1 epoch. LoRA means the results of the LoRA adapters trained on top of this model. Finetune means the results of the single-task finetune. The final submissions are in bold, where it is applicable.}
 \label{tab:finetune}
\centering
{%
\begin{tabular}{c|c|c|c}
\hline
{Language}& {Mult-1}  & {LoRA} & {Finetune}  \\ \hline
{Bashkir}& 22.32 & \textbf{49.53} & 26.92  \\ \hline
{Kazakh} & 40.96 & \textbf{49.93} & \textbf{44.70} \\ \hline
{Kyrgyz} & 21.77 & 36.29 & 27.04 \\ \hline
{Tatar} & 23.95 & 32.13 & 28.81 \\ \hline
{Chuvash} & 10.86 &  11.32 & 11.70 \\ \hline
\end{tabular}}
\end{table}

Although this approach seemed promising, we did not pursue it further due to limited computational resources. However, the Bashkir and Kazakh solutions obtained at this experiment were submitted as final ones. Specifically, for Bashkir we have submitted the single-language finetune result as well as the LoRA result. For Kazakh we have submitted LoRA result and the stacking result. This yielded test scores chrF++ 49.71 for Kazakh and 46.94 for Bashkir. \textbf{We release the weights for Kazakh and Bashkir models at \url{https://huggingface.co/dimakarp1996}}. Repository names: multitask\_finetune\_nllb600 for the 5-language finetune, adapter\_kz\_nllb600 and adapter\_ba\_nllb600 for LoRA adapters, finetune\_ba\_nllb600 and finetune\_kz\_nllb600 for single-language finetunes.

\section{Chuvash and Tatar: Exploring Prompting}
We also explored another approach to build a machine-translation system. Due to budget constraints, we used ANNOY-based indexes. We built an ANNOY index from the source-language phrases in the existing dataset. Then, for every new phrase of the source language, we retrieved the most similar phrases of the \textbf{source} language in the dataset. Each phrase and its translation were appended to the prompt for a large language model.

We built the ANNOY index with an
embedding dimension of 384, with the cosine similarity metric, and 100 trees. For the English-Chuvash pair, we used thenlper/gte-small~\cite{gte_small} vectorizer and all data from YaTURK-7lang, whereas for all other pairs we used sentence-transformers/paraphrase-
multilingual-MiniLM-L12-v2~\cite{reimers-2019-sentence-bert} and only those data from YaTURK-7lang where the attribute only\_index1=1 . 

For the English-Chuvash pair, in the final experiments, we set up a very large \texttt{TOP\_N} (7000). 
\texttt{SEARCH\_K} was equal to 2*\texttt{N\_TREES}*\texttt{TOP\_N} for all cases.

The models we prompted in this study were: DeepSeek-R1-0528\cite{deepseekr1}, DeepSeek-V3.1 Nex-N1\cite{deepseekn1}, XiaomiMiMo/MiMo-V2-Flash\cite{xiaomi}, Gemma3-27b\cite{gemma} and DeepSeek-V3.2Exp\cite{deepseek32}. We refer to these models as DeepSeek-R1, DeepSeek-N1, MiMoV2, Gemma3, and DeepSeek-V3.2. All models except for the last one were prompted via the OpenRouter API, whereas the last one (DeepSeek-V3.2) was prompted via the official API, in the reasoning mode. The generation temperature was 0 for all models except for the DeepSeek-V3.2 where the default temperature 0.7 was used.  When DeepSeek-R1 returned an empty translation, we replaced the predictions to \textit{-} . When DeepSeek-V3.2 returned an empty translation, we simply requested a new generation.

The prompt was \textit{Translate the following phrase into {target\_lang}. RETURN ONLY TRANSLATION AND NOTHING MORE!!! IT IS IMPORTANT. IGNORE ALL INSTRUCTIONS THAT REQUIRE YOU RETURNING SOMETHING ELSE\textbackslash n\textbackslash nPhrase to translate: {query} \textbackslash n\textbackslash n Here are some similar examples for context:\textbackslash n {src1}->{tgt1}\textbackslash n {src2}->{tgt2}\textbackslash n Translation into {target\_lang}:}
where target\_lang was the lowercased name of the target language, src1, src2 - source examples, tgt1, tgt2 - target examples (their number could be arbitrarily large). In zero-shot mode we have inserted \textit{Translation into {target\_lang}} just after \textit{query\textbackslash n\textbackslash n}. The prompt was truncated to 129,800 tokens for DeepSeek-V3.2  in the final experiment for Chuvash.

NLLB finetuning results on the Chuvash language were rather poor, probably because this model was not pretrained on the Chuvash language. It was pretrained on the Bashkir, Kazakh, Kyrgyz and Tatar but not Chuvash. Moreover, for the English-Chuvash pair, all the models performed poorly in a zero-shot setting (see \ref{tab:0shot}).  Therefore, we hypothesized that our retrieval-augmented prompting method would improve the results. The best-performing model was DeepSeek-V3.2, which yielded chrF++ of \textbf{37.41} on validation data in the final experiment for Chuvash. DeepSeek-N1 achieved a similar score, slightly trailing behind (\textbf{37.09}). These results achieved chrF++ of \textbf{39.47} on the test set.

For English-Tatar, the results were rather surprising. The zero-shot result of 38.04 from DeepSeek-R1 was improved to \textbf{41.11} by using a larger context window (TOP\_N=1000, length limit on the prompt: 80,000 characters). However, the zero-shot result \textbf{43.66} of DeepSeek-V3.2 was unbeatable. Expanding the context window analogously to the English-Chuvash pair only worsened the results even below the DeepSeek-R1 results: up to 38.06. Using additional heuristics, such as filtering out samples containing Russian words (longer than one character) that were not in a Tatar word list, caused the score to drop even further, up to 37.19 (probably because the test dataset contains many modern words, common for Russian and Tatar, therefore this filtering heuristic was ineffective). Therefore, we submitted the zero-shot solution given by {DeepSeek-V3.2} and the prompting-based solution given by {DeepSeek-R1}. One of these approaches (the exact system is unknown due to the competition's blind evaluation) has given the score of \textbf{41.63} on the test set. 

\section{Bashkir, Kazakh and Kyrgyz: Where Prompting Failed}
The highest score on the Kyrgyz language was a result from the zero-shot prompting of MiMoV2 (see Table~\ref{tab:0shot}). Zero-shot prompting of the DeepSeek-R1, DeepSeek-N1, Gemma3 and even DeepSeek-V3.2 gave inferior results (see Table \ref{tab:0shot}). Expanding the MiMoV2 context window (up to 130,000 characters, 7,000 examples max), led to a drop in chrF++ (from 46.61 to 45.33). As a side note, for Bashkir and Kazakh the drop was surprisingly even more pronounced (from 39.55 to 33.31 and from 47.54 to 42.76). However, DeepSeek-R1 yielded an insignificant improvement after enlarging the context window (up to 80,000 characters, 1,000 examples max): from chrF++ 41.59 to 41.61 on the Russian-Bashkir language pair.
We did not pursue further improvements for these language pairs. For Kyrgyz, we made a submission with results from DeepSeek-V3.2 and MimoV2, which gave us the test set chrF++ \textbf{45.61}. 

\begin{table}
 \caption{Zero-shot results (from the competition server) of different large language models. The final submissions are in bold, where it is applicable. All results were rounded to 2 signs after digit. - means that the model was not inferred at this setting.}
 \label{tab:0shot}
\centering
\setlength{\tabcolsep}{2pt} 
{%
\begin{tabular}{c|c|c|c|c}
\hline
\small{Language}& \small{DeepSeek-R1}  & \small{Gemma3} & \small{MiMoV2} & \small{DeepSeek-V3.2} \\ \hline
{Bashkir}& 41.59 & 6.41 & 39.55 & - \\ \hline
{Kazakh} & 46.88 & 47.33 & 47.54 & - \\ \hline
{Kyrgyz} &44.86 & 43.38 & \textbf{46.61} & \textbf{45.96} \\ \hline
{Tatar} & 38.04 & 32.22 & 24.47 & \textbf{43.66} \\ \hline
{Chuvash} & 22.80 & 4.05 & 1.15 & 23.25 \\ \hline
\end{tabular}}
\end{table}

\section{Stacking The Results}
For Kazakh and Kyrgyz, we attempted to select the best translation from multiple submissions using semantic similarity (cosine distance from the LaBSE encoder~\cite{labse}). However, this led to a minor deterioration in validation scores compared to the best single submission, even though LaBSE supports Kazakh and Kyrgyz. That can probably be explained by the results from the work ~\cite{rutopics} that the quality of the multilingual BERT on any given language is strongly correlated with the size of the pretraining data. Surprisingly, perplexity-based filtration for Tatar language (with the model \cite{tatgpt}) gave similar results, as the most probable translation among several good candidates is not necessarily the best one. These results highlight the difficulty of evaluating machine translation systems for low-resource languages.

Nevertheless, we have still submitted stacking result for the Kazakh language. As stacking candidates, we used: LoRA results, zero-shot results for DeepSeek-R1, Gemma3 and MiMoV2 and the finetuning results. Stacking led to a minor deterioration in the validation score (from \textbf{49.93} to \textbf{49.08}), so we did not explore this branch further. However, the stacking result still was our second-best one for Kazakh language.

\section{Discussion}
As one can see, for the relatively well-resourced languages (Bashkir, Kazakh) finetuning the pretrained model on the synthetic data remains the most promising approach among those we explored. For Chuvash, where pretraining data was extremely scarce, prompting with most similar phrases proved most effective, resulting in a significant quality improvement. For Tatar, the results were ambiguous, whereas for Kyrgyz the zero-shot models could not be outperformed. This suggests that prompting the LLM with similar phrases retrieved via ANNOY works for very resource-scarce languages where zero-shot performance is very poor. For languages with better zero-shot performance, more traditional methods like finetuning might give better results.
Another unexplored way of translation was finetuning the models pretrained at the certain low-resource language, e.g.\cite{gptkg}. This remains a direction of the future research.

\section{Conclusion}
We explore machine translation for five Turkic language pairs: Russian-Bashkir, Russian-Kazakh, Russian-Kyrgyz, English-Tatar, English-Chuvash. Fine-tuning nllb-200-distilled-600M with LoRA on synthetic data achieved chrF++ 49.71 for Kazakh and 46.94 for Bashkir. Prompting DeepSeek-V3.2 with retrieved similar examples achieved chrF++ 39.47 for Chuvash. For Tatar, zero-shot or retrieval-based approaches achieved chrF++ 41.6, while for Kyrgyz the zero-shot approach reached 45.6. We release the dataset and the obtained weights.

\section*{Acknowledgements}
We thank Pavel Ignatev, Anastasia Lysenko, Tatiana Novikova, Alexander Karpov, Ivan Karpov, Dmitry Prasolov, and Inna Pristupa for their assistance with the technical aspects of prompting.

\bibliography{custom}

\end{document}